\documentclass[10pt,twocolumn,letterpaper]{article}

\usepackage{iccv}
\usepackage{times}
\usepackage{epsfig}
\usepackage{graphicx}
\usepackage{amsmath}
\usepackage{amssymb}
\usepackage{comment,hhline}
\usepackage{float}
\usepackage{times}
\usepackage{epsfig}
\usepackage{graphicx}
\usepackage{amsmath}
\usepackage{amssymb}
\usepackage{gensymb}
\usepackage{enumitem}
\usepackage{multirow}

\usepackage[pagebackref=true,breaklinks=true,letterpaper=true,colorlinks,bookmarks=false]{hyperref}

\iccvfinalcopy 

\newcommand{\argmax}{\operatornamewithlimits{argmax}}
\newcommand{\argmin}{\operatornamewithlimits{argmin}}
\newtheorem{thm}{Theorem}
\everymath{\displaystyle}

\ificcvfinal\fi
\begin{document}

\title{Camera Calibration by Global Constraints on the Motion of Silhouettes}

\author{Gil Ben-Artzi \\
 The Faculty of Mathematics and Computer Science\\ 
 Weizmann Institute Of Science, Israel
 }

\maketitle

\begin{abstract}
We address the problem of epipolar geometry using the motion of silhouettes. Such methods match epipolar lines or frontier points across views, which are then used as the set of putative correspondences. We introduce an approach that improves by two orders of magnitude the performance over state-of-the-art methods, by significantly reducing the number of outliers in the putative matching. We model the frontier points' correspondence problem as constrained flow optimization, requiring small differences between their coordinates over consecutive frames. Our approach is formulated as a Linear Integer Program and we show that due to the nature of our problem, it can be solved efficiently in an iterative manner. Our method was validated on four standard datasets providing accurate calibrations across very different viewpoints.  
\end{abstract}

\section{Introduction}

Multi-camera systems are becoming increasingly more popular for $3D$ reconstruction, marker-less motion capture, surveillance, and even for "smart homes". Traditionally,  epipolar geometry is computed by finding the corresponding points between cameras. However, in such a setting many camera pairs are from very different viewpoints and consequently, not enough reliable feature points can be matched automatically. In cases where the silhouettes of the objects in the scene are available or are easily extracted, it has been proposed to use their motion for calibration \cite{sinha2004camera}. These methods match tangent epipolar lines on the convex hull of the silhouettes across views \cite{sinha2010camera}. For each frame, a similarity function for every possible matching of the corresponding tangent epipolar lines is evaluated and the pair with the highest similarity is selected as a putative correspondence \cite{Ben-Artzi_2016_CVPR}. Recovering the corresponding tangent epipolar lines induces matching between the projections of special points, denoted as \emph{extremal frontier points}, across views where occlusions do not occur (see \cite{furukawa2004structure,sinha2010camera, cipolla1995motion,lazebnik2007projective}  for a detailed description). Hereafter, when referring to frontier points, we will refer to the extremal $3D$ frontier points or to the image points that represent their projections. The epipolar geometry is evaluated by using RANSAC \cite{fischler1981random} with the putative list of matching epipolar lines or frontier points.

\begin{figure}[t]
\begin{center}
\includegraphics[width=0.48\linewidth]{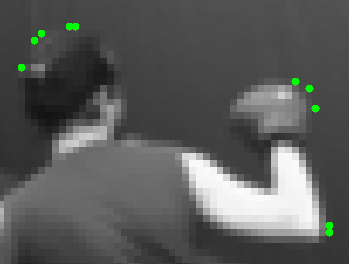}
\includegraphics[width=0.48\linewidth]{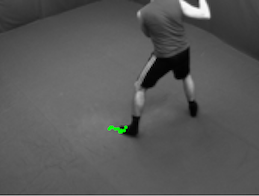}\\
(a)~~~~~~~~~~~~~~~~~~~~~~~~~~~~~~~~~(b)
\end{center}
\caption{(a) The set of possible frontier points are denoted by green circles. It is difficult to distinguish between true and false frontier points because often the candidates are at nearby positions and share the same tangents to the convex hull. (b) The trajectory of the true frontier points over $15$ consecutive frames. We look for frontier points whose positions vary slowly over time.}
\label{fig:intro}
\end{figure}

A key limitation of previous approaches is the high number of outliers in the recovered list of corresponding epipolar lines or frontier points. This is mainly because two nearby points on the convex hull might make only a minor difference in their associated tangents. Here, we present an approach that markedly reduces the number of outliers. This leads to an improvement of two orders of magnitude in the performance. Unlike previous approaches where the putative matching is carried out for each frame independently, we require that the positions of frontier points in consecutive frames will vary slowly. Fig.~\ref{fig:intro}.a shows a set of possible frontier points in the image denoted by full green circles. Since nearby points share similar tangents, it is hard to recover the true frontier points. Fig.~\ref{fig:intro}.b shows the trajectories of the true frontier points over $15$ frames. By looking for frontier points that vary slowly over time, we can accurately recover their positions, up to sub-pixel accuracy.

We model the problem as a constrained flow optimization problem. At each time instant we look for two corresponding pairs of frontier points. This translates into finding two paths in the graph under a set of constraints. The graph is formulated as an integer programming (IP) problem whose global maximum can be recovered with reasonable complexity. However, since there is a large number of variables and constraints, the solution to this problem might be slow and not scalable for long sequences. We therefore provide an efficient algorithm for obtaining an approximate solution by iteratively applying the shortest path algorithm. We show that for any practical camera setting, the global optimum is recovered.

This paper therefore contributes by presenting: (a) a reformulation of the problem of finding frontier points across views as a constrained graph optimization problem, (b) a practical algorithm to recover its global maximum, and (c) an accurate calibration method with superior performance over state-of-the-art methods. We validated our approach on several standard datasets and show that it is highly effective.

\section{Related Work}
\label{sec:prior}

The most common uses of silhouettes in multi-camera systems are for shape-from-silhouettes \cite{cheung2003shape,forbes2006shape,aganj2007spatio} and camera calibration \cite{hernandez2007silhouette,sinha2010camera,boyer2006using,ramanathan2000silhouette,zhang2009self}. In shape-from-silhouettes, the goal is to recover the visual hull \cite{laurentini1994visual,miller2006exact} of the object. In calibration, it is assumed that the motion of the silhouette is fully observed across different views.  Correspondences are established between frontier points across different views based on the epipolar tangency constraints \cite{cipolla1995motion} and are then used to compute the epipolar geometry. Most methods require a specific configuration that cannot be applied in a general setting, which is considered here. These include calibrated cameras, static objects, orthographic projection models, or known (turntable) motion  \cite{franco2003exact,furukawa2006robust,wong2001structure,mendonca2001epipolar,hernandez2007silhouette}. Calibration can be carried out without explicitly finding the corresponding tangent epipolar lines, such as in \cite{boyer2006using, yamazoe2006multiple}. These methods require a good initial guess to converge. 

In this paper we consider calibration in the most general setting where only the motion of silhouettes is available without further assumptions. Sinha and Pollefeys \cite{sinha2010camera} considered such a setting for calibration. They searched for the epipoles by randomly sampling lines from the tangent envelope of the silhouette.  They used RANSAC for extracting the most plausible solution.  Ben-Artzi et al.\cite{Ben-Artzi_2016_CVPR} proposed a temporal binary descriptor, denoted as a motion-barcode,  for suggesting the corresponding epipolar lines across views. They sampled the lines from the tangent envelope according to the similarity induced by their descriptor. Importantly, they showed that accuracy and runtime are markedly improved. In both methods there is a significant number of outliers and the frontier points in the current frame are matched without taking into account the previous or the next corresponding points. Kasten et al.\cite{Kasten2016} used the same descriptor for calibrating crowded scenes. However, they considered noisy, low-resolution images and provided limited accuracy.   

We used the similarity proposed by the binary motion-barcode descriptor \cite{Ben-Artzi_2016_CVPR} as one of the cues for our algorithm. Similar motion-based binary descriptors have been proposed and used by \cite{ermis2010activity},\cite{drouin2010camera}.  Both methods assume a planar structure of the scene. Ben-Artzi et al. \cite{benevent} used a similar descriptor for matching events across different views, but their method does not provide accurate localization. Pundik and Moses~\cite{pundik2010video} introduced a similar motion-based descriptor,
line signal, and used it for video  synchronization. It depends on the color and was used under the assumption of known calibration. 

\begin{figure}[t]
\begin{center}
 \includegraphics[width=1\linewidth]{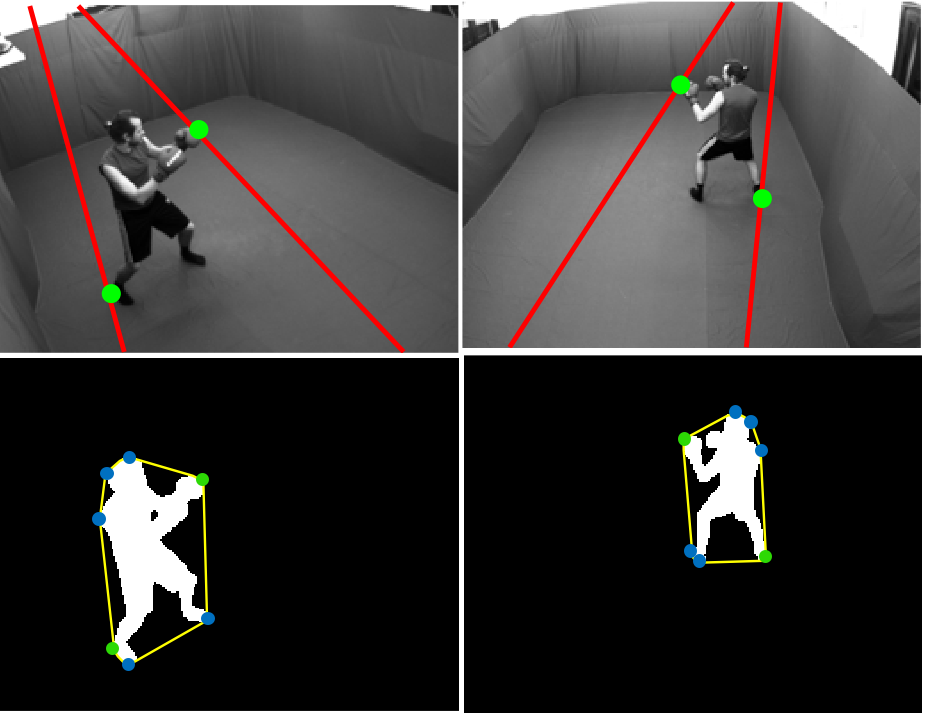}\\
\end{center}
\caption{(a) Top row. Two corresponding images captured from different views. The red lines are the epipolar lines. The green circles are the ground truth frontier points. (b) Bottom row. The frontier points can only be derived from a small set of points.  These are points on the convex hull intersecting with the silhouette; they are denoted by a circle and are termed \emph{critical points}. The green critical points are also the ground truth frontier points.}
\label{fig:matching}
\end{figure}

\section{Our Model}

We assume that we have two sequences of binary silhouette images, each captured from a different view. Consider the image captured from the first view as the left image and the image captured from the other view as the right image. In each of the images, the convex hull of the silhouette is extracted and the subset of points on the intersections of the convex hull and the silhouettes are identified. These points are termed \emph{critical points}. The true correspondence of the extremal frontier points across views can only be found between critical points \cite{cipolla2000visual,sinha2010camera}, such as is illustrated in Fig.~\ref{fig:matching}. In addition, we assume that we are given a similarity measure for the correspondence of critical points across views. 

We model the matching problem as a constrained flow in a graph. Our goal is to recover the two paths that maximize the flow. Each path is a sequence of corresponding pairs of frontier points across views, at each time instant.

\subsection{Formulation}

\begin{figure}[t]
\begin{center}
 \includegraphics[width=1\linewidth]{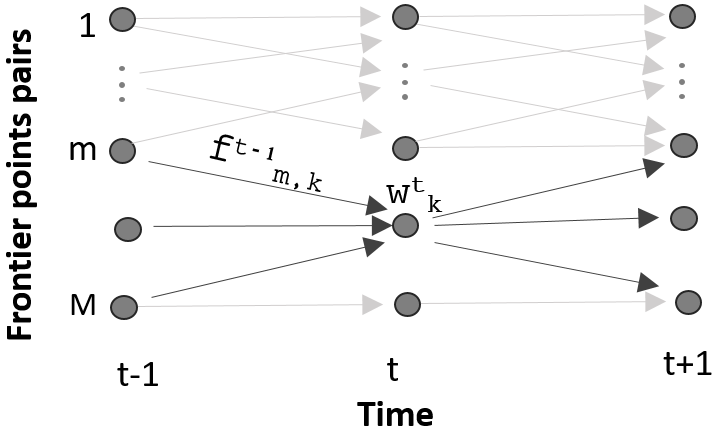}\\
\end{center}
\caption{Our problem is represented as a directed acyclic graph (DAG). Every vertex in graph $v^k_t$ is associated with a binary variable $w^t_k$ and every edge in graph $e^{t-1}_{m,k} $ is associated with a binary variable $f^{t-1}_{m,k}$. The vertices represent all possible pairs of matched frontier points across views.}
\label{fig:graph}
\end{figure}

We introduce a directed acyclic graph (DAG) $G=(V,E)$. Let $x^t_i$ denote a critical point in the left image and $x'^t_j$ denote a critical point in the right image, both at time instant $t$. Each vertex $v^t_k$ in the graph represents a pair of two critical points $(x^t_i,x'^t_j)$. Each edge between vertices represents an admissible transition between a true match at time $t$ and a true match at time $t+1$. In our case we consider all possible transitions between the time instants as admissible transitions. The number of vertices at each time instant is $M=K_1 \times K_2$ vertices, where $K_1$ and $K_2$ are the total number of critical points in the left and right images, respectively. Let $N(k) \subset \{1...K\}$ be the neighborhood of vertex $k$. There is an edge $e^t_{i,j}$ from $v^t_i$ to $v^{t+1}_j$ if and only if $j \in N(i)$.  Every vertex is associated with a binary variable $w^t_i \in \{0,1\}$, indicating whether the current pair of critical points are a true match of frontier points. Each edge $e^t_{i,j}$ is associated with a binary variable $f^t_{i,j}$, indicating whether both $v^t_i$ and  $v^{t+1}_j$ are a true match. Such a graph is illustrated in Fig~.\ref{fig:graph}.

We now define the set of constraints that will ensure that the set of flows accurately represent our matching problem.  First, for each vertex in the graph, the flow conservation implies that the sum of flows coming into a vertex at time $t-1$ equals the sum of flows outgoing from the vertex at time $t$:
\begin{equation}
\forall t,j \sum_{i \in N(j)} f^{t-1}_{i,j}=\sum_{k \in N(j)} f^t_{j,k} = w^t_j.
\label{eq:1}
\end{equation}

We refer to $N(x)$ as either edges outgoing from vertex $x$ or incoming into vertex $x$, depending on the meaning of the index $x$.

Second, in our formulation, each vertex is either a true or false match between frontier points. This implies that at most, one unit of flow leaves the vertex:
\begin{equation}
\forall t,i \sum_{j\in N(i)} f^t_{i,j} \leq 1.
\label{eq:2}
\end{equation}

Third, the flows must be positive:
\begin{equation}
\forall t,i,j f^t_{i,j} \geq 0.
\label{eq:3}
\end{equation}

Fourth, we are looking for two paths in the graph. To enforce this, we introduce the source vertex $v_{src}$, which is linked to all vertices of the first time instant and the target vertex $v_{trg}$, which is linked to all vertices of the last time instant. The flow is from the source vertex to the target vertex. The source vertex generates exactly two units of flows and the target vertex absorbs exactly two units of flow:
\begin{equation}
\sum_{j \in N(src) } f_{src,j} = 2, \sum_{j \in N(trg) } f_{j,trg} = 2.
\label{eq:4}
\end{equation}

For brevity we do not explicitly write the flow conservation constraints for these vertices.


Fifth and last, the minimal distance between the frontier points can often be bounded. Consequently, we look for two paths in the graph that are $C$ pixels far from each other.  This constant can be adjusted according to the specific setting, although in practice one fixed constant is sufficient for all camera pairs over all datasets, see Section \ref{sec:constant}. For a given vertex $v^t_i=(x,x')$, let $D(i) \subset \{1...M\}$ be the nearby vertices at the same time instant:
$$
D(i)=\big\{j| v^t_j=(y,y'), min \{ d(x,y),d(x',y') \} < C \big\},
$$
where $d$ is the Euclidean distance and $C$ is the constant representing the required distance between the frontier points. The last constraint is therefore:
\begin{equation}
\forall t,k \sum_{j \in N(k) } f^t_{k,j} + \sum_{m \in D(k)} \sum_{n \in N(m)} f^t_{m,n} \leq 1.
\label{eq:5}
\end{equation}

\subsection{Linear Integer Program}

We assume that we have an estimator for the probability that a given pair of critical points across views is the corresponding frontier points:
\begin{equation}
p^t_i=P(w^t_i=1).
\label{eq:estimator1}
\end{equation}

In addition, let us further assume that we can also estimate the conditional probability of a true match in a vertex given a true match in a predecessor vertex:
\begin{equation}
p^{t-1}_{i,j}=P(w^t_j=1|w^{t-1}_i=1).
\label{eq:estimator2}
\end{equation}

Our objective is to obtain a set of binary assignments ${\bf w}$ that can best explain our estimations:
\begin{equation}
{\bf w}^{*}= \argmax_{{\bf w} \in {\bf S}} P(\bf{w}) ,
\label{eq:obj}
\end{equation}
where ${\bf S}$ is the space of feasible solutions satisfying constraints \eqref{eq:1}-\eqref{eq:5}. Assuming that our model obeys the Markov property, objective \eqref{eq:obj} can be written as:
\begin{align}
{\bf w}^{*} & =  \argmax_{{\bf w} \in {\bf S}} \sum_{t,i} log P(w^t_i) + \sum_{t,j,i\in N(j)} log P(w^t_j |w^{t-1}_i).
\label{eq:obj_sum}
\end{align}

Because $w^t_i$ is a binary variable, we can write 

\begin{align}
log P(w^t_i)& = w^t_i log P(w^t_i=1)+(1-w_i^t) log P(w_i^t=0) \nonumber \\
&= w_i^t log(\frac{p^t_i}{1-p^t_i})+J(p^t_i),
\label{eq:first_term}
\end{align}
where $J(p^t_i)$ is a term that does not depend on $w^t_i$. Similarly, 
\begin{equation}
log P(w^t_j |w^{t-1}_i) \propto f^{t-1}_{i,j} log ( \frac{p^{t-1}_{i,j}}{1-p^{t-1}_{i,j}}).
\label{eq:second_term}
\end{equation}

Plugging Eqs.~\eqref{eq:first_term}-\eqref{eq:second_term} into Eq.~\eqref{eq:obj_sum}, ignoring the terms that do not depend on $\bf{w}$, and expressing $w^i_t$ in terms of flows, we obtain the following Integer Program:

\begin{align}
\underset{\bf{f}}{\text{maximize}} & ~~~~\sum_{t,i}  log(\frac{p^t_i}{1-p^t_i}) \sum_{j\in N(i)} f^t_{i,j}+ \label{eq:integerp} \\
& \sum_{t,j,i\in N(j)} log ( \frac{p^{t-1}_{i,j}}{1-p^{t-1}_{i,j}}) f^{t-1}_{i,j}  \nonumber \\
\text{subject to}
&~ \forall t,i,j ~ f^t_{i,j} \geq 0 , f^t_{i,j} \in \{0,1\}  \nonumber \\
& ~\forall t,i \sum_{j\in N(i)} f^t_{i,j} \leq 1 , \nonumber \\
& ~\forall t,j \sum_{i\in N(j)} f^{t-1}_{i,j} -\sum_{k \in N(j)} f^t_{j,k} = 0 \nonumber \\
& ~\sum_{j \in N(src) } f_{src,j} = 2\;, \sum_{j \in N(trg) } f_{j,trg} = 2 \nonumber\\
& ~\forall t,k \sum_{j \in N(k) } f^t_{k,j} + \sum_{m \in D(k)} \sum_{n \in N(m)} f^t_{m,n} \leq 1 \nonumber 
\end{align}

\subsection{Optimization}

Using standard LP solvers, a solution can be found for our Integer Program~\eqref{eq:integerp}. Because solving IP is NP-complete, finding such an exact solution is not feasible for real-life applications. One can relax the problem into a continuous Linear Program and obtain a solution at a polynomial time, but the constraints matrix of ~\eqref{eq:integerp} is not Totally Unimodular \cite{hoffman2010integral} and it is not likely to converge to the original optimal solution. However, we show that in our case, the optimal integer solution can always be computed in a much better way than the brute-force approach. This is stated in the following theorem:
\begin{thm}
The optimal integer solution of~\eqref{eq:integerp} can be recovered in $O(TK^4)$, where $T$ is the number of time instants and $K$ is the maximum number of vertices at each time instant.
\end{thm}

The proof is given in the supplementary material. However, such an approach can only be applied for moderately sized problems and is not scalable. Therefore, in the following we show how to compute the optimal solution in an efficient way that can also be applied to large size problems. We show that except for degenerated cases, the solution is the optimal one. It has been applied successfully for all camera pairs over all the datasets.

Similarly to \cite{suurballe1974disjoint,bhandari1999survivable}, we iteratively use the shortest path algorithm to solve the problem. This approach is also discussed in \cite{iqbaldisjoint}.

We construct a directed acyclic graph (DAG) $G'=(V',E')$ with the same structure as the graph $G$. An edge $e'^t_{i,j}$ is assigned the weight:
\begin{equation}
u(e'^t_{i,j})= -log(\frac{p^{t+1}_j}{1-p^{t+1}_j})  - log ( \frac{p^{t}_{i,j}}{1-p^{t}_{i,j}}).
\label{eq:weights}
\end{equation}

The weights for the edges outgoing from the source vertex are assigned only the first term from~\eqref{eq:weights}, and the weights for the edges incoming to the target vertex are set to zero. The optimal solution to our Integer Program $\bf{f}^*$ can be written on the graph $G'$ as:
$$
	{\bf f}^*=\argmin_{f\in S} \sum_{t,i,j \in N(i)} u(e'^t_{i,j}) f^t_{i,j},
$$

where $S$ is the set of feasible solutions of \eqref{eq:integerp}, satisfying the constraints given in Eqs.~\eqref{eq:1}-\eqref{eq:5}. To recover the optimal solution, we use the following:
\begin{itemize}
\item Find the shortest path $s_1$ on $G'$.
\item For each vertex in the shortest path $v'^t_i \in s_1$, set the outgoing edges of the vertex and the nearby vertices $\{v' | v' \in D(i)\}$ to $\infty$. 
\item Find the new shortest path $s_2$ on the modified graph.
\item Return $\hat{f}= \{s_1 \cup s_2 \}$.
\end{itemize}

The above procedure is much more efficient than recovering the optimal integer solution and it can be implemented easily. In our case, we have a trellis graph and the shortest path is computed by dynamic programming \cite{bellman1952theory}. Assuming there are at most $K$ vertices at each time instant and $T$ time instants, the solution is at a cost of $O(TK^2)$, instead of $O(TK^4)$ for the optimal integer solution. The key drawback is that it does not always guarantee that the global maximum can be recovered. However, as stated in the following theorem, unless there is a degenerate configuration, the solution will always be converged to the global one.

\begin{thm}
Let $f^*=\{s^*_1 \cup s^*_2\}$ be the optimal two-path solution for the Integer Program~\eqref{eq:integerp} and let $C$ be the constant selected for constraint ~\eqref{eq:5}. Then, $f^* \neq \hat{f}$ if and only if (a) $s_1^* \neq s_1$ and $s_2^*\neq s_2$ and (b) $ \exists t,i,j  ~~\textrm{s.t.}~~  v^t_i \in s_1^*, v^t_j \in s_2^* ~~\textrm{and}~~ d(v^t_i,v^t_j)<2C $.
\end{thm}

The proof for the above theorem is given in the supplementary material. 
The outcome is that the optimal solution will not be recovered only if the two pairs of frontier points at the same time instant will be less than $2C$ pixels apart. In practice, one fixed constant is sufficient for all the camera pairs over all datasets (see Section \ref{sec:constant}). 

\begin{table}[tb]
\begin{center}
\begin{tabular}{ |l|l|l|l|l|}
\hline 
         & Boxer &  Girl  & Street & Kung-Fu   \\
     \hline  \# Pairs & 6 &       28 &   15 & 300  \\
      \# Frames  & 778 &       200 &   250 & 200   \\
      Type  & Real &       Real &  Real & Graphics  \\
\hline
\end{tabular}
\end{center}
\caption{ Dataset properties
\label{table:datasetProperties}}
\end{table}

\begin{figure}[tb]
\begin{center}
   (a)  \includegraphics[width=0.3\linewidth]{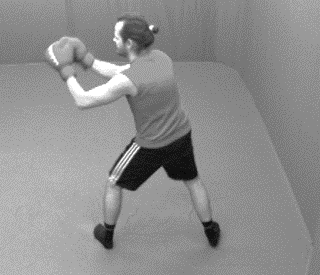} 
   (b) \includegraphics[width=0.3\linewidth]{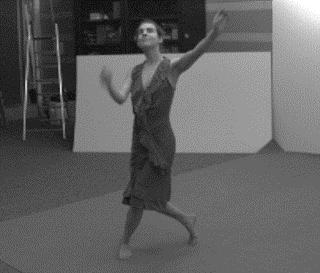} \\
 (c)    \includegraphics[width=0.3\linewidth]{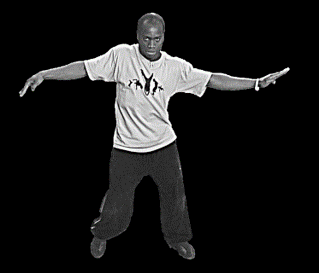} 
(d)     \includegraphics[width=0.3\linewidth]{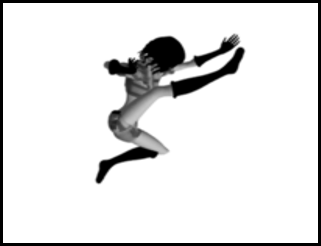} 
\end{center}
\caption{The datasets. (a) Boxer. (b) Girl. (c) Street (Dancer). (d) Kung-Fu. }
\label{fig:datasets}
\end{figure}

\section{Experiments}
\label{sec:results}

We compared our method with the state-of-the-art approaches of Ben-Artzi et al.\cite{Ben-Artzi_2016_CVPR} and Sinha et al.\cite{sinha2010camera}. For accurate comparisons, we followed the exact same procedures as in Ben-Artzi et al.\cite{Ben-Artzi_2016_CVPR}. The fundamental matrix $F$ was computed by RANSAC-based sampling, based on the putative points’ correspondences.  It is optimized using the same non-linear Levenberg-Marquardt (LM) optimization procedure.

\subsection{Datasets}

We used four datasets. The datasets are Boxer \cite{Boxer}, Girl \cite{DancingGirl}, Street (Dancer) \cite{StreetDancer}, and Kung-Fu \cite{kungfu}.  Table ~\ref{table:datasetProperties} presents the properties of each dataset and Fig.~\ref{fig:datasets} shows sample images.  All datasets used are publicly available, along with ground truth calibrations.

\subsection{Implementation Details}

We implemented both approaches in MATLAB using standard libraries on a computer with i7 quad-core CPU and 8GB memory. The Linear Integer Program was formulated using CVX \cite{cvx} and solved with MOSEK solver~\cite{mosek}.  We found the shortest path in our trellis by dynamic programming \cite{bellman1952theory} and in all experiments, we report the results of the fast iterative algorithm. 

As our input, we used two sources of information. The first is the coordinates of the critical points in each image, computed by the intersection of the convex hull of each silhouette. The second is a similarity measure of the correspondence between two critical points across views. This was used to produce  $p^t_i, p^t_{i,j}$ of Eqs. ~\eqref{eq:estimator1}-\eqref{eq:estimator2}, which will be described next.

\subsubsection{The Conditional Probability Estimator}

We constructed $p^t_{i,j}$, based on the distance between the two pairs. Let $v^t_i=(x,x')$ be the pair of critical points with coordinate $x$ in one image and $x'$ in the other image, at time instant $t$. Let $v^{t+1}_j=(y,y')$ be the pair of critical points at time instant $t+1$. The conditional probability estimator is according to the assumption that the Euclidean distance between the coordinates is a random variable distributed normally with zero mean and unit variance: 
$$
d([x,y]-[x',y']) \sim \mathcal{N}(0,1), 
$$
where $\mathcal{N}(\cdot)$ is the normal distribution and $[\cdot,\cdot]$ denotes concatenations of vectors. 

\subsubsection{The Similarity Estimator}

Assume that we have access to a similarity measure $sim(l,l')$, such that it gives us an estimate of how likely these two lines $l,l'$ are corresponding epipolar lines. Assume that $v^t_k$ represents the critical pair $(x,x')$. For $v^t_k$ to represent a true correspondence between frontier points across views, the corresponding epipolar lines must be incident to these points. Since we do not know the exact tangent line, for each critical point there is a set of possible tangent lines. This is the set of lines $L=\{l_i\}_{i=1}^K$ from the \emph{tangent envelope} of the silhouette at time instant $t$, which are incident to the critical point $x$. Similarly, consider $L'$ for $x'$. We define the similarity estimator as:
$$
p^t_i \propto max( \{ sim(l,l') | l \in L, l' \in L'\}). 
$$

\begin{figure}[t]
\begin{center}
 \includegraphics[width=1\linewidth,height=0.6\linewidth]{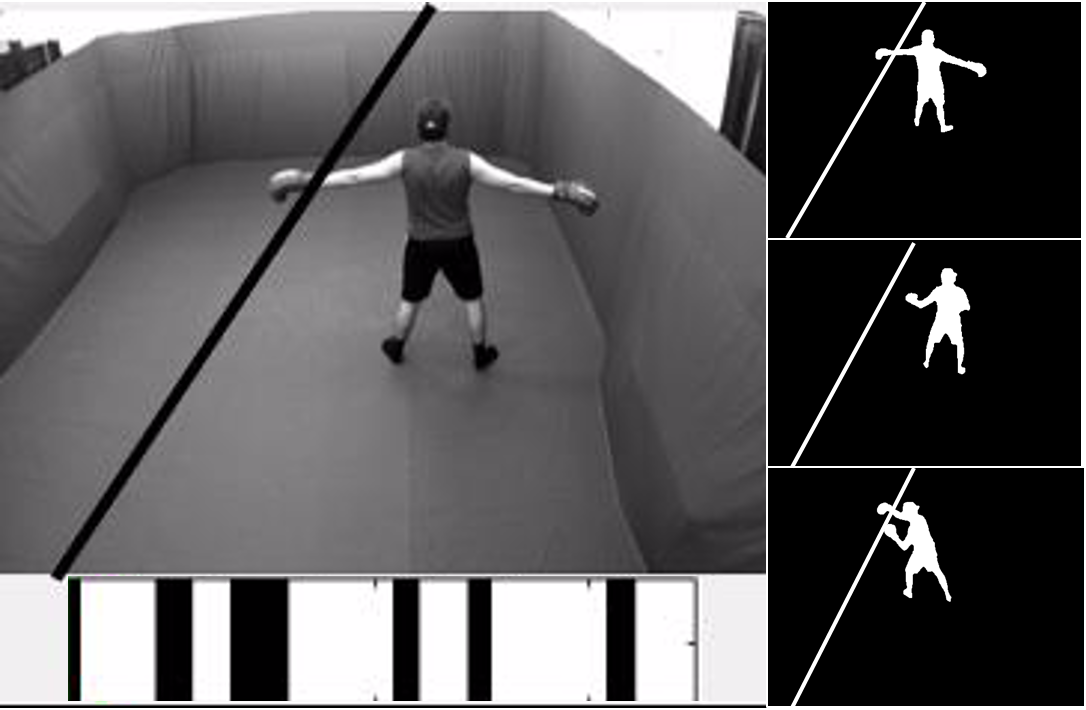}\\
 ~~~~~~~~~~~~~~~~(a)~~~~~~~~~~~~~~~~~~~~~~~~~~~~~~~~~~~~~~~~~~(b)
\end{center}
\caption{The Motion Barcode descriptor used for the similarity of epipolar lines. (a) The image of the boxer and a selected (arbitrary) line. Over time, at each instant the intersection of the line with the silhouette is tested and recorded as zero or one, accordingly. The recorded series of zero/one is illustrated at the bottom of the image, where black is one and white is zero. (b) Three different silhouettes at different time instants. In the first and the last time instants the motion barcode is recorded as one and for the second time instant it is recorded as zero; thus it is $[1,0,1]$.}
\label{fig:motionbarcode}
\end{figure}

\subsubsection{Motion-Based Similarity of Epipolar Lines} 

Here we briefly describe the similarity measure for two lines to represent the corresponding epipolar lines. It is the input to our similarity estimator. It was used in \cite{Ben-Artzi_2016_CVPR,Kasten2016,benevent}, which also provides a detailed description.

The similarity measure is based on a descriptor denoted as \emph{motion barcode}. For a given image line, a motion barcode is constructed as follows. For each image in the sequence, the intersection of the line with the silhouette is tested. The value of the motion barcode at that time instant is set to either zero or one, accordingly. Thus, it is a binary sequence of the same length as the number of images in the sequence. It has been shown that if two lines are indeed corresponding epipolar lines, their motion barcodes should be very similar \cite{Ben-Artzi_2016_CVPR}. The similarity between the motion barcodes of two lines is the correlation between the binary sequences.  The motion barcode is illustrated in Fig.~\ref{fig:motionbarcode}.

\subsubsection{The Graph}
\label{sec:constant}

We set the required minimal distance between the frontier points $C$ for $D(\cdot)$ to $15$ pixels, in all the experiments across all datasets and camera pairs, without fine-tuning it for each camera pair individually. The required distance depends on the specific setting and can in principle be adjusted for each specific case accordingly. We verified that the same single constant is valid for all valid frontier points over all the frames in the datasets.

Our method results in two paths in the graph. Each path consists of a set of vertices representing a match between frontier points across views. The flow conservation constraints required selecting vertices at each frame, even if the similarity estimator and the conditional probability estimator output have very low probabilities. Similarly to \cite{Ben-Artzi_2016_CVPR}, we used a threshold of $0.95$ on the motion barcodes’ similarity measure to remove unreliable matches whose associated similarity is lower than this threshold.  

\begin{figure}
\begin{center}
\includegraphics[width=0.72\linewidth]{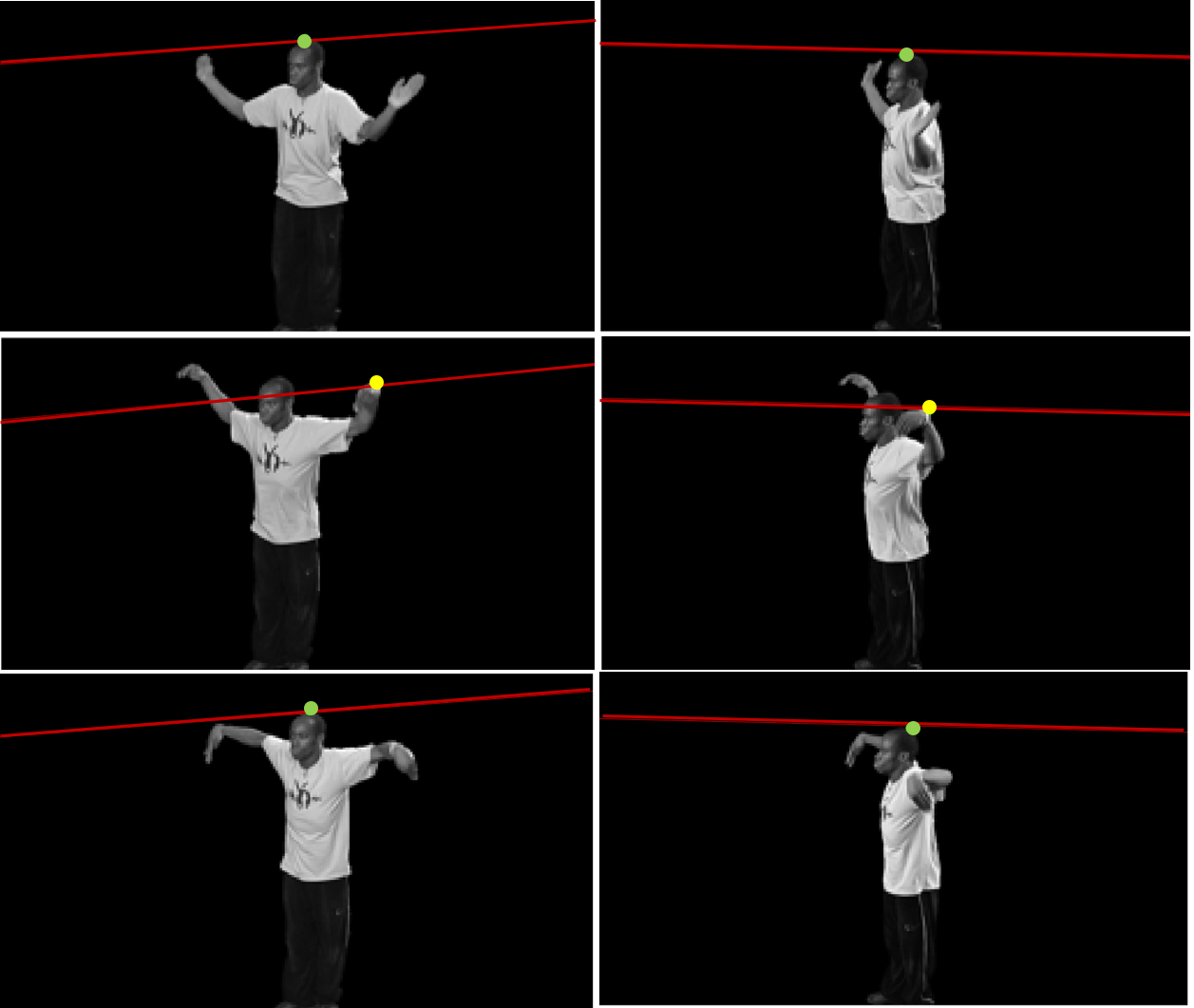} 
\end{center}
   \caption{True correspondences of non-frontier points. On rare occasions, less than 1\%, our method also matches non-frontier points.  Each row is two views from the same time instant, ordered by time from the top downward. The red lines are the epipolar lines. The first and last rows from the top show matched frontier points, denoted in green. The middle row is the time instant with non-frontier points that are matched, denoted in yellow. These are true corresponding critical points. See text for more details. }
\label{fig:reg}
\end{figure}

In addition to recovering frontier points, on rare occasions our method also matches non-frontier points.  Fig.~\ref{fig:reg} shows a time instant for which non-frontier points are matched. It presents samples from eight consecutive frames, ordered by time from the top downward; each row shows corresponding views for the same time instant. The red lines denote the corresponding epipolar lines. The actor raises his hands and then lowers them. At the time instances represented by the first and last rows from the top, the true frontier point correspondences are recovered, denoted by green circles. At the time instant represented by the second row from the top, the matched points denoted by yellow circles are indeed true corresponding points but they are not frontier points. These cases are less than 1\% of the matches and occur when a) the similarity estimator outputs for these critical points which are non-frontier points very high probability to be a true correspondence and b) the corresponding true frontier points are matched in nearby positions, immediately before and after this specific time instant. Due to the regularization, the non-frontier points are preferred over the frontier points.  

\subsection{Evaluation}

\subsubsection{Metrics}

{\bf Quality of Fundamental Matrix}. A standard metric for assessing the quality of the fundamental matrix is the symmetric epipolar error \cite{hartleyMVG} defined as:
$$
Q(F)=\frac{1}{N} \sum_i d(x_i',F x_i)^2 + d(x_i,F^T x'_i)^2,
$$

where $F$ is the evaluated fundamental matrix and $\{(x_i,x_i')\}_{i=1}^N$ are the corresponding points across views. \\

\noindent
{\bf Efficiency}. The efficiency measure is the expected number of RANSAC iterations needed to achieve a given or a better quality of a fundamental matrix (a lower error).

\subsubsection{Methodology}
\noindent
{\bf Efficiency Comparison}. For the generation of the efficiency metric, we followed the protocol by \cite{Ben-Artzi_2016_CVPR}. They selected the best error every 1000 iterations and optimized it using non-linear Levenberg-Marquardt (LM) optimization procedure. For example, in the Girl dataset there are $28$ camera pairs and therefore, our sample size is $2800$ errors. We then calculated the cumulative distribution function (CDF) of the error and used it in the evaluation as in the baselines. \\

\noindent
{\bf Inliers' Probabilities}. The efficiency metric used by \cite{Ben-Artzi_2016_CVPR} is based on (a) the probability of having an inlier, (b) the selection of the best sample using the ground truth points, and (c) the non-linear optimization technique. Thus, it might not directly reflect the quality of the putative correspondences using the tested approach. We therefore present the probability of having an inlier for various thresholds using our approach, which can be used for future reference. It is calculated as follows. We measured the symmetric epipolar distance for each recovered corresponding pair of points using our approach with respect to the ground truth fundamental matrices. Using this distance, the fraction of inliers can be evaluated for each dataset and required accuracy.\\

\noindent
{\bf Overhead}. Our method requires an additional step of finding the two paths on the graph. The runtime cost is equivalent to $300-1200$ RANSAC iterations, depending on the length of the sequence and the number of vertices. It was added to the comparisons.\\

\begin{figure}[tb]
\begin{center}
   \includegraphics[width=0.75\linewidth]{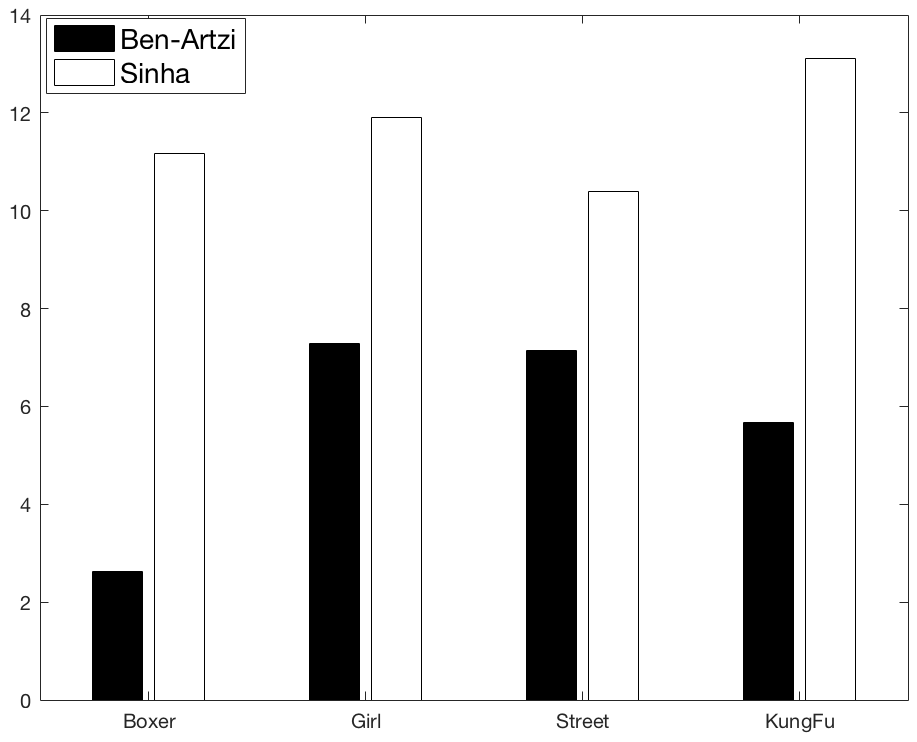} \\
 \end{center}
\caption{ 
The ratio between the expected number of RANSAC iterations required to reach a given accuracy of the fundamental matrix, using our method and the baselines. Accuracy is measured using the symmetric epipolar distance with respect to the ground-truth points. The best hypothesis is selected every 1000 RANSAC iterations and is optimized using non-linear Levenberg-Marquardt (LM) method. The ratio is over all accuracy for each dataset. The x-axis is the dataset and the y-axis is the $log2$ mean ratio.
\label{fig:ratio}}
\end{figure}

\begin{table}[tb]
 \footnotesize
\begin{center}
\begin{tabular}{ |l|c|c|c|c|c| }  
\hline
&  Boxer & Girl  &  Street &  Kung-Fu & Mean\\ \hline
 Sinha &  2211.2  & 447.52 & 354.54 & 186.11 & 799.84 \\
Ben-Artzi &  8.86 & 10.38 & 20.85 & 5.56  & 11.41 \\ \hline
\end{tabular}
\end{center}
\caption{The ratio, as in Fig.~\ref{fig:ratio}, measured for accuracy of 0.8 for each dataset and with respect to each method. The last column is the mean over all datasets. 
\label{table:ratio_exact}}
\end{table}



\subsection{Results}

\noindent
Fig.~\ref{fig:ratio} shows our main results. It presents the ratio between the expected number of RANSAC iterations per accuracy using our method and the baselines. The best hypothesis is selected every 1000 RANSAC iterations and is optimized using the non-linear (LM) method.  The accuracy is the symmetric  epipolar  distance  with  respect to ground-truth points. The ratio is the mean over all accuracy for each dataset. The x-axis is the dataset name and the y-axis is the $log2$ mean ratio. Our approach introduces an improvement of orders of magnitude, in all datasets and for all accuracy levels. Overall, the mean improvement is $92.82$ with respect to \cite{Ben-Artzi_2016_CVPR} and $993.71$ with respect to \cite{sinha2010camera}.  Table~\ref{table:ratio_exact} presents the same ratio for a required accuracy level of $0.8$, measured for each dataset over all camera pairs.  For such an accuracy, the mean ratio is $11.41$ with respect to \cite{Ben-Artzi_2016_CVPR} and $799.84$ with respect to \cite{sinha2010camera}. Table~\ref{table:samples} presents detailed comparisons, as was done in \cite{Ben-Artzi_2016_CVPR}. A very high accuracy can be reached very quickly. For example, reaching an accuracy of $0.3$ for the Street dataset requires on average only $6245$ RANSAC iterations. Generally, the higher the required accuracy, the more the improvement can be introduced. Our approach can reach in a reasonable number of iterations an accuracy level of $0.2$.

\begin{table}[tb]
\scriptsize
\begin{center}
\begin{tabular}{|l|l|r|r|r|r|r|r|}
\hline
\multicolumn{2}{|c|}{ Epipolar Distance } & 1 & 0.8 & 0.5 & 0.4 & 0.3 & 0.2\\ 
\hline
\multirow{2}{*}{Boxer} 
& Sinha   &2.9M&2.9M&-&-&- & -    \\   
& Ben-Artzi   &5K&12K&111K  &996K  &- & -    \\     
& Ours  &1230&1354&13K &300K&600K& - \\
\hline
Girl    & Sinha & 149K&388K&13M&-&- & -  \\
 & Ben-Artzi & 4K&9K&129K&918K&13.7M & -  \\   
 & Ours &828&867&1195&4562&30283& 560K   \\
\hline

Street & Sinha     & 159K&340K&1.8M&7.4M&- & -    \\
 & Ben-Artzi     & 7K&20K&255K&616K&1.2M & -    \\
  & Ours & 928&959&1279&2142&6245& 62.5K\\
\hline

\multirow{2}{*}{Kung-Fu} 
& Sinha  &65K&134K&822K&1.9M&8.6M& - \\
& Ben-Artzi  &2K&4K&23K&71K&302K& - \\
& Ours &711&720&814&998&2058& 13.4K\\
\hline

\end{tabular}
 \end{center}
\caption{  The expected number of RANSAC iterations required to reach a given accuracy of the fundamental matrix, using our approach and the baselines. In each dataset, the number of iterations is averaged over all cameras pairs.  Accuracy is measured using the symmetric epipolar distance with respect to the ground-truth points. The best hypothesis is selected every 1000 RANSAC iterations and is optimized using non-linear Levenberg-Marquardt (LM) method.  Empty cells indicate that the required accuracy was not attained.
\label{table:samples}}
\end{table}

\begin{table}[tb]
 \footnotesize
\begin{center}
\begin{tabular}{ l|c|c|c|c|c|c| }    
Epipolar Distance  & 1 & 0.8 & 0.5 & 0.4 & 0.3 & 0.2\\ \hline
Boxer  &  0.37 & 0.30 &   0.19 & 0.15 &  0.11 & 0.07  \\ 
Girl   &  0.46 & 0.38 &   0.24 & 0.19 & 0.15 & 0.11    \\ 
Street  &  0.66 & 0.53 &   0.35 & 0.28 & 0.21 & 0.14     \\ 
Kung-Fu&  0.65 & 0.56 &   0.39 & 0.32 & 0.25 & 0.17    \\ 
\end{tabular}
\end{center}
\caption{The inliers’ probabilities for each required accuracy and for each dataset, averaged over all camera pairs in the dataset. It was measured by determining the symmetric epipolar distance of our recovered points with respect to the ground truth fundamental matrices.
\label{table:inliersFP}}
\end{table}

\begin{table}[tb]
 \footnotesize
\begin{center}
\begin{tabular}{ r|c|c|c|c|c|c| }
Epipolar Distance&1&0.8&0.5&0.4&0.3&0.2\\
\hline Mean Inliers prob.&0.53&0.44&0.29&0.23&0.18&0.12\\
Required Samples&390&1440&26K&135K&752K&1.28M\\ 
\end{tabular}
\end{center}
\caption{ First row. The mean probability of having an inlier of a given accuracy for all camera pairs over all datasets. Second row. The required number of samples using the RANSAC procedure for reaching the required accuracy. On average, only 26K samples are needed to reach a sub-pixel accuracy of $0.5$. 
\label{table:inliersAverage}}
\end{table}

Table~\ref{table:inliersFP} presents the inliers' probabilities. It was computed based on the symmetric epipolar distance to the ground truth’s fundamental matrices for each camera pair in the dataset. Table~\ref{table:inliersAverage} shows the mean probability of having an inlier of a required accuracy for all camera pairs over all datasets. It also shows the required number of RANSAC samples needed to reach each accuracy. On average, only $1440$ samples are needed to reach a sub-pixel accuracy of $0.8$. For a required accuracy of $0.5$, only $26K$ samples are needed on average.  Table~\ref{table:numPts} shows the number of recovered putative corresponding pairs for a given accuracy using our approach, based on the inlier probabilities. For the boxer dataset, we were able to recover more than $100$ pairs with an accuracy of $0.5$ and $60$ pairs with an accuracy of $0.3$.  

\begin{table}[tb]
 \footnotesize
\begin{center}
\begin{tabular}{ c|c|c|c|c| }    
Accuracy \textbackslash Dataset &  Boxer & Girl  &  Street &  Kung-Fu \\ \hline
Any  &  547 & 108 & 127 & 150 \\
0.5 &  104 & 26 & 44 & 58 \\
0.3 &  60 & 16 & 27 & 37 \\
\end{tabular}
\end{center}
\caption{The number of corresponding points recovered for each dataset, over all the camera pairs. The first row is the total number of points recovered. The second row is the number of inliers having a symmetric epipolar distance equal or less than $0.5$, and the third row is for a symmetric epipolar distance of $0.3$.
\label{table:numPts}}
\end{table}

\begin{figure}[tb]
\begin{center}
\includegraphics[width=0.75\linewidth,height=0.55\linewidth]{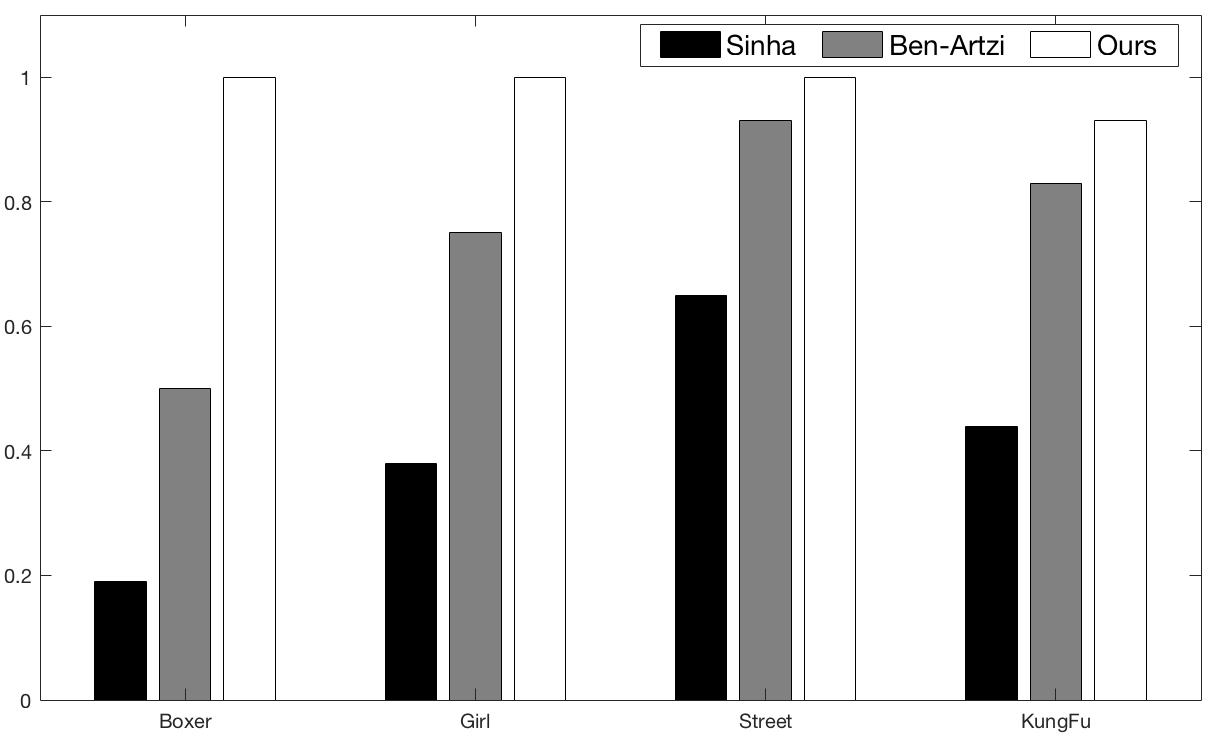} 
  \end{center}
\caption{The percentage of cameras that converged for a required accuracy of $0.8$. Using our approach, all camera pairs, in all datasets except Kung-Fu converged. For the Kung-Fu dataset, $88\%$ of the cameras converged. See text for more details.}
\label{fig:converged}
\end{figure}


Fig.~\ref{fig:converged} shows the percentage of converged cameras using our method and the baselines, for a required accuracy of $0.8$. In all datasets except Kung-Fu, using our approach, all camera pairs converged. In the Kung-Fu dataset, however, $88\%$ of the camera pairs converged. This is due to the fact that there are camera pairs for which the epipoles are inside the convex hull. In such cases, tangent-based methods often fail to recover accurate matching points. These cases of failure are inherent in such approaches, see \cite{Ben-Artzi_2016_CVPR} for an illustrative example. Nevertheless, due to the dynamic nature of the object there are often frames within the same sequence where the epipoles are also outside the convex hull, which is sufficient for calibration. For example, for a required accuracy of $1.5$, only two camera pairs failed in the Kung-Fu dataset and all other camera pairs over all the other datasets converged.

\section{Conclusion}
\label{sec:conclusion}
We introduced a graphical model for calibrating a multi-camera system from the motion of silhouettes. Our approach recovers corresponding points efficiently and accurately, outperforming state-of-the-art methods by several orders of magnitude. It is optimized very efficiently, providing a practical solution. 
Our approach fits seamlessly into a silhouettes-based pipeline, and it can be used automatically each time a new sequence of silhouettes is captured.
{\small
\bibliographystyle{ieee}
\bibliography{egbib}
}

\end{document}